# A ROS2 based communication architecture for control in collaborative and intelligent automation systems

Endre Erős*[a], Martin Dahl[a], Kristofer Bengtsson[a], Atieh Hanna[b], Petter Falkman[a]

[a]*Department of Electrical Engineering, Chalmers University of Technology, Gothenburg 41296, Sweden*
[b]*Department for Research and Technology Development, Volvo Group Trucks Operations, Gothenburg 41755, Sweden*

**Abstract**

Collaborative robots are becoming part of intelligent automation systems in modern industry. Development and control of such systems differs from traditional automation methods and consequently leads to new challenges. Thankfully, Robot Operating System (ROS) provides a communication platform and a vast variety of tools and utilities that can aid that development. However, it is hard to use ROS in large-scale automation systems due to communication issues in a distributed setup, hence the development of ROS2. In this paper, a ROS2 based communication architecture is presented together with an industrial use-case of a collaborative and intelligent automation system.

## 1. Introduction

The ever-increasing demand for product variability and quality, as well as the need for shorter production and delivery time, introduces new challenges in production. Some of these challenges can be solved by using autonomous and collaborative robots [1]. To ease the integration and development of collaborative and intelligent systems, various platforms have emerged as middle-ware solutions where the Robot Operating System (ROS) [2] stands out.

---

* Corresponding author. Tel.: +46733292237.
  *E-mail address:* endree@chalmers.se

Since the incarnation of ROS in 2007 [3] by Willow Garage, the number of ROS users has seen a major increase with more than 16 million total downloads in the year 2018 [4]. Currently, ROS2 [5] is being developed, where the communication layer is based on Data Distribution Service (DDS) [6] to enable large scale distributed control architectures. As it is presented in this paper, this improvement paves the way for use of ROS2-based architectures in real-world industrial automation systems.

Since ROS2 is behind ROS when it comes to the number of packages and active developers, communication bridges are used to pass messages between ROS and ROS2. These bridges allow the community to utilize strengths of both ROS and ROS2 in the same system, i.e. to have an extensive set of developed packages and at the same time have a robust way to communicate between machines. This is crucial in modern industry because it enables robust integration of state-of-the-art tools and algorithms necessary for control in collaborative and intelligent automation systems. Since these large-scale automation systems are usually distributed in nature, they require a well structured and reliable communication architecture. Therefore, this paper presents variations of a ROS2 based communication architecture with these traits.

While enabling integration and communication is greatly beneficial, it is just one part of the challenge. The overall control architecture also needs to plan and coordinate all actions of robots, humans and other devices as well as to keep track of everything. Mixed human-machine industrial environments where operations are carried out either collaboratively, coactively or individually, demand non-traditional control strategies. A few ROS based frameworks that try to handle these control strategies are ROSPlan [7], SkiROS [8], eTaSL/eTC [9] and CoSTAR [10]. These frameworks are however mainly focused on single robot systems and as such lack the infrastructure to support large scale automation systems. Intelligent and collaborative systems often comprise of several robots, machines, smart tools, human-machine interfaces, cameras, safety sensors, etc.

A use-case in this paper illustrates the challenges of achieving flexible automation in an industrial setup and highlights the need to use ROS due to great integration with smart devices and algorithms. A brief overview of the communication layer in ROS and ROS2 is given in Section 2. A description of the use-case is given in Section 3 and the design of the communication architecture is described in Section 5. Sequence Planner [11], a state-of-art task planner and discrete controller is shortly explained in Section 4 and the paper is concluded in Section 6.

## 2. The Robot Operating System (ROS) and ROS2

ROS is a collection of tools and libraries that aids robotics software development [3]. Software is written in form of nodes that exchange messages using a transport layer called TCPROS [12] based on standard TCP/IP sockets. ROS has a centralized network configuration which means that there has to exist a running ROS master [13]. The master takes care of naming and registration services so that other ROS nodes could find each other on the network and communicate in a peer-to-peer fashion. Having a configuration where all nodes depend on a central ROS master is not robust, especially if nodes in a network are distributed on several computers.

To improve on this design limitation, developers of ROS2 implemented Object Management Group's standard DDS [14] as the communication middleware considering it to be scalable, robust and well-proven in mission-critical systems [15]. As a result, DDS enables implementation of ROS in industrial use-cases.

There is a number of available implementations of DDS, allowing ROS2 users to choose an implementation from a vendor that suits their needs. This is done through a ROS Middleware Interface (RMW) [16], which also exposes Quality of Service (QoS) policies. ROS2 users have the ability to choose between options of QoS policies for supported RMW implementations, which enables them to further modify the transfer layer.

There are implementations which offer a more direct access to the real-time publish-subscribe (RTPS) wire transfer protocol. While not fulfilling the full DDS API, they still provide sufficient functionality for ROS2. An example implementation is eProsima's FastRTPS [21] and it is the implementation of choice for the setup described in this paper. Default QoS settings are kept and they specify reliability, volatile durability and KEEP_LAST history. An extensive but outdated study compares performances of ROS middlewares [22]. In contrast to the mentioned study, we found that FastRTPS performed without problems.

Since ROS2 uses DDS to exchange messages, the Discovery Module of the underlying RTPS protocol enables ROS2 nodes to automatically find each other on the network [14, 15], thus there is no need for a ROS2 master.

44*Endre Erős, Martin Dahl, Kristofer Bengtsson, Atieh Hanna, Petter Falkman / Procedia Manufacturing 00 (2019) 000–000*   3Setting up a stable ROS multi-master system before ROS2 was introduced was rather cumbersome, and it often resulted in using other communication platforms such as Kafka or ZeroMQ for exchanging messages between different ROS systems. Several specialized ROS packages that enable communication in multiple ROS master systems already exist, e.g., *multimaster_fkie* [13], *rocon* [17], *ZeroMQ-ROS* [18] and *Kafka-ROS Bridge* [19].

Currently, there are still cases where there is a need to use old ROS packages when designing new systems. In distributed cases, it is possible to use the communication layer capabilities of ROS2 together with local ROS systems. Utilizing ROS2 and several ROS masters enables us to partition a ROS system into local, single computer hubs, where a hub includes a ROS master and one or multiple local nodes. This is be exemplified with an industrial use-case and described in more detail.

## 3. An industrial use-case of collaborative and intelligent automation

Benefits of utilizing both ROS and ROS2 are evident in a collaborative robot assembly station in a Volvo Trucks engine manufacturing facility located in Skövde, Sweden [20]. The aim of this use-case is to present possibilities in designing a workstation where both humans and robots work in a collaborative or coactive fashion. Particularly in this use-case, a collaborative robot and a human operator should perform assembly operations on a diesel engine, Fig. 1(a).

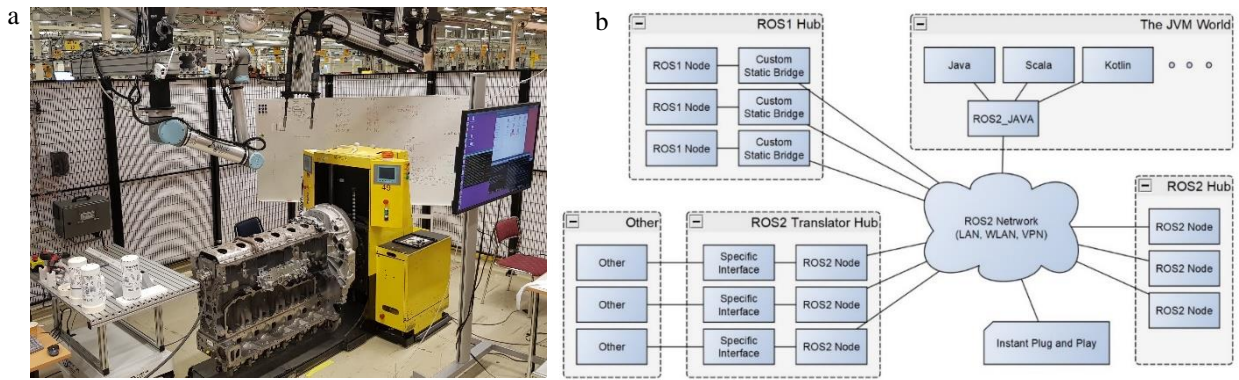

Fig. 1: (a) A Collaborative Robot Assembly Station; (b) High-level communication overview

The physical setup of this assembly station consists of a six Degree of Freedom (DoF) collaborative robot, an autonomous mobile platform, two different specialized end-effectors, a smart tool that can be used as an end-effector, a docking station for the end-effectors, a lifting system and docking station for the smart tool, a camera and RFID reader system and eight computers dedicated for different tasks. The collaborative assembly process is briefly explained in the next part of this section.

An Automated Guided Vehicle (AGV) carrying a diesel truck engine and an autonomous mobile platform (MiR100) carrying kitted material to be assembled on an engine enter the Collaborative Robot Assembly Station. A ladder frame, three oil filters and several oil transport pipes are to be mounted on the engine in this station. Before the collaborative execution of these operations can begin, an authorized operator has to be verified with a RFID reader. After verification, the operator is greeted by the system and instructions for tasks that the human should perform are shown on a screen. If no operator is verified, some operations can still be executed independently by the robot, however, violation of safety zones around the robot trigger a safeguard stop.

In the case of collaborative assembly, a dedicated camera system keeps track of the human assuring safe coexistence. During positioning of the AGV and the MiR100 in the assembly station, the Universal Robots (UR10) robot attaches to a special end-effector needed for manipulation ladder frame manipulation. Since the ladder frame is quite heavy, the robot and the human collaborate in transporting it from the kitted MiR100 to the engine.

After placing the ladder frame on the engine, the operator informs the control system with a button press on a smart watch, after which the UR10 leaves the current end-effector and attaches itself to a smart tool used for fastening bolts. During this tool change and as it was instructed on the instructions screen, the operator is placing twelve pairs of bolts needed for assembling the ladder frame on the engine and indicates completion of this task through the smart watch. Now, the control system knows that bolts are in place and the UR10 starts the tightening operation with the smart fastener tool. Since tightening of these bolts takes some time, the operator can simultaneously mount three oil filters on the engine. If the robot finishes the tightening operations first, it leaves the smart tool in a floating position above the engine and waits for the operator in an idle home position.

The operator uses the smart watch again to let the system know that the oil filters are in place. This event makes the robot attach to the third end-effector and start performing the oil filter tightening operations. During the same time, the operator attaches two oil transport pipes on the engine, and uses the same smart tool that the robot has left floating to tighten plates that hold the pipes to the engine. After executing these operations, the AGV with the assembled engine and the empty MiR100 leave the Collaborative Robot Assembly Station.

From the use-case example described in this section, it is evident that there is a need to model, analyze and control large scale intelligent automation systems in a smart way. A video showing the described use-case in action can be found here: https://youtu.be/YLZzBfY7pbA

## 4. Sequence Planner as a Discrete Event Controller

Sequence Planner (SP) [11] is a tool for modeling, analyzing and control of automation systems. In SP, automation systems are modeled using operations and variables. It includes algorithms for a variety of use cases related to modeling, for instance synthesizing control logic, verification and visualizing complex operation sequences in different projections. SP uses a hierarchical control approach based on planning of low level *abilities*.

Abilities are operating directly on control state which is estimated by applying *transformation pipelines* to different ROS2 topics. Because ROS messages are strictly typed, pipelines can be generated automatically to populate a SP model with variables that correspond to fields in the message types used. More complex pipelines can also be used, for instance for aggregating messages into a single state variable in SP or for applying some discretization function to reduce the state space of the controller.

Modeling in SP is done by composing abilities (modeled independently of each other) using specifications that disable undesired interactions between them. By disabling a minimum of undesired behavior, a lot of flexibility is left for on-line planning algorithms to find the currently most suitable desired behavior. This is done by employing synthesis based on supervisory control theory [23]. Modeling examples can be found in [24] and [25].

On top of abilities, desired behavior is modeled using *operations*. Operations are matched to sequences of underlying abilities based on on-line planning. This allows for a clear separation of two control layers, i.e. operations do not need to know about underlying detailed abilities.

Planning is continuously performed in a receding horizon fashion, allowing the control to be intelligent about which abilities to execute based on current state of the system. Fig. 2(a) shows SP during simultaneous execution of four operations (in green) where live resource usage is visualized at the bottom. In order to assure transparent and robust message transfer between Sequence Planner and the rest of the system, a well organized communication architecture has to be designed.

## 5. Communication Architecture

The architecture described here focuses on ROS hubs which allow us to separate the system into functional entities. Each of these hubs exchanges messages of a specific type with the rest of the system, where the types are defined for the function that the entity has. It is enough to define and build a message type if a hub consists only ROS2 nodes. In the case of ROS hubs, network bridges have to be used which transfer ROS messages to ROS2 and vice-versa. In the use-case described in this paper, Sequence Planner is exchanging messages with the system using the Java client library for ROS2. Fig.1(b) gives a high-level overview of the hubs in a ROS2 system.

*5.1. ROS Hubs*

The system is split into hubs where a hub can be considered an independent group of at least one ROS or/and ROS2 node. One or more hubs can be present on one computer, but a single ROS hub shouldn't spread on more than one computer. ROS hubs are considered to have an accompanying set of bridging nodes which pass messages between ROS and ROS2. Hubs which consist of merely ROS2 nodes can be considered ROS2 hubs.

In this use-case, ROS nodes are distributed on several computers including several single-board computers (SBC's). Beside ROS2 hubs, independent ROS hubs which are necessary for this use-case are located on some of the computers. Table 1 gives an overview of used computers and functions of the ROS hubs on each of them.

Table 1. Setup component and role overview.

| No. | Name    | ROS v. | Computer | OS        | Arch. | Network   | Explanation                           |
|-----|---------|--------|----------|-----------|-------|-----------|---------------------------------------|
| 1   | Tool ECU| Cr+Me  | Rasp. Pi | Ubuntu 18 | ARM   | LAN1+VPN  | Smart tool and lifting system control |
| 2   | RSP ECU | Crystal| Rasp. Pi | Ubuntu 18 | ARM   | LAN1+VPN  | Pneumatic conn. control and tool state|
| 3   | Dock ECU| Crystal| Rasp. Pi | Ubuntu 18 | ARM   | LAN1+VPN  | Gives states of docked end-effectors  |
| 4   | MiRCOM  | Crystal| LP Alpha | Ubuntu 18 | amd64 | LAN2+VPN  | ROS2 (VPN) to/from REST (LAN2)        |
| 5   | MiR     | Kinetic| Intel NUC| Ubuntu 16 | amd64 | LAN2      | Out-of-the-box MiR100 ROS Suite       |
| 6   | RFIDCAM | Crystal| Desktop  | Win 10    | amd64 | LAN1+VPN  | Publishes RFID and Camera data        |
| 7   | UR10    | Cr+Kin | Desktop  | Ubuntu 16 | amd64 | LAN1+VPN  | UR10 ROS Suite                        |
| 8   | DECS    | Crystal| Laptop   | Ubuntu 18 | amd64 | LAN1+VPN  | Sequence Planner on JVM               |

The smart tool and lifting system control algorithm were independently developed for ROS by another team of developers and implemented on a Raspberry Pi with an added piggyback electronic control unit (ECU). Using ROS2 with the network bridges, it was easy to integrate this independent ROS hub into the system.

The autonomous platform MiR100 from Mobile Industrial Robots comes with an embedded Intel NUC computer that is running all the necessary ROS control nodes locally. This comes as a factory default feature with a REST API provided for communication. A ROS2 node runs on a SBC onboard the MiR100 that serves as a translator hub via REST on the MiR100's LAN. This SBC is also included in a virtual private network (VPN) using a 4G communication dongle, and as such, it is able to receive ROS2 messages from SP in the same VPN and pass them on as REST commands. This way, the MiR100's factory default ROS system stays isolated and unchanged.

ROS hubs are part of a VPN, which enables convenient remote access for debugging or data collection. Moreover, an added benefit of a VPN setup it the ensured safety of the system.

Communication in ROS and ROS2 is message based, meaning that nodes that act as publishers or subscribers exchange messages between themselves. Nodes from hubs in this system communicate with SP using custom message types designed for this use-case. These hubs communicate with the rest of the system via bridges since computers running them are equipped with ROS2 as well.

*5.2. Dynamic vs. Static bridges*

The ROS2 package *ros1_bridge* [26] provides a network bridge which enables passing messages between ROS and ROS2. The dynamic bridge can automatically open bridges while listening to topics from both sides. According to the argument passed while starting the dynamic bridge, messages can be passed in some direction, or both.

Since inconsistency of received messages can close the dynamic bridge and lead to a temporary loss of transfer, static bridges are used instead. These bridges are modified so that they could pass custom messages of a single topic in one direction only. Static bridges are more customizable and their performance doesn't depend on the periodical consistency of the messages.

These static bridges are opened on each ROS hub; they are essentially scripts that start both a ROS and a ROS2 node. These scripts provide a way to specify the direction of the bridge, which topic should be bridged and which type of message is flowing on that topic.

For example, to bridge a ROS node from a hub that is subscribed to Command messages and publishes State messages, one would have to open two static bridges. The one-directional static bridge for the Command topic would contain a ROS2 Subscriber node and a ROS Publisher node and would pass Command messages to the node of the hub. For the other way, a one-directional static bridge for the State topic would contain a ROS Subscriber node and a ROS2 Publisher node and would pass State messages from the node of the hub to others.

*5.3. Messages and Communication*

A set of packages exists that defines common message and service types for ROS and ROS2. This set covers a wide range of types. Still, some setups call for designing custom messages whose usability extends those of common types. In the setup described here, two families of messages were compiled and used to communicate between SP and every other ROS or/and ROS2 hub in the system.

Table 2. Example Command and State messages for Robot Mover.

| Command | | | State | | |
|---|---|---|---|---|---|
| Type | Name | Content | Type | Name | Content |
| string | action | MOVEJ | string | robot_name | TARS |
| string | robot_type | UR10 | bool | fresh_msg | TRUE |
| string | robot_name | TARS | float32 | t_plus | 0.8 |
| string | pose_type | JOINT | bool | got_reset | FALSE |
| string | pose_name | HOME | string[ ] | error_list | [ ] |
| float32 | speed_scal | 0.1 | command | echo | Command |
| float32 | acc_scal | 0.2 | bool | moving | FALSE |
| float32 | goal_toll | 0.01 | float32 | actual_pose | POSE13 |

One family of messages contains custom command types which are used to send state based commands to the hubs, while the other family contains custom state messages that the hubs use to send out data about their state. Example messages from each family are shown in Table 2. It should be noted that the Echo message is nested in the State message and as such is part of it. The purpose of the Echo message is to inform SP that the intended node has received the message.

Another interesting thing to note is that this message type can also be used to communicate with the MiR100 platform as well. It is enough to specify the correct robot type and name, and to update the action to match the capabilities of the robot. An additional example of a used message type is shown in Table 3. This message is used to save a current pose of a robot and can also be applied to any robot in the system.

Table 3. Example Command and State messages for Pose Saver.

| Command | | | State | | |
|---|---|---|---|---|---|
| Type | Name | Content | Type | Name | Content |
| string | action | UPDATE | string | robot_name | TARS |
| string | robot_type | UR10 | bool | fresh_msg | FALSE |
| string | robot_name | TARS | float32 | t_plus | 38 |
| string | pose_type | JOINT | command | echo | command |
| string | pose_name | HOME | string | done_action | updated |

Dividing message types into these two families is of course not necessary, one could pass any messages between hubs. Still, in larger systems, it is convenient to compile such types of messages and limit the amount of topics that is bridged since designing and debugging can be simplified. Also, having state based communication in a ROS system, enables the nodes to stay stateless which can simplify the system and make the handshaking more convenient. These messages are continuously sent out from both sides with a constant frequency, utilizing the decoupled nature of a publish-subscribe transfer. This way, the communication in the system is made even more resilient.

## 5.4. Hub vs. Node type oriented communication architecture

There are several ways to organize nodes, hubs, messages and topics in a distributed ROS system. Designing the communication architecture in the described collaborative robot assembly station was approached with considerations of possible future expansions. One possible expansion of the system would be the inclusion of other robots from other vendors, thus nodes and hubs are designed with the ability to handle equipment agnostic messages. This leads to a primarily node type oriented communication architecture since families of handler nodes could be copied on different hubs to handle different equipment while receiving messages of same type.

In a system where hubs differ in structure, it is valid to pursue a more hub oriented communication architecture. This means that each hub is looked at as a unique set of nodes that are communicating with the rest of the system. This implies that, in order to structure the communication in a large multi-hub system where hubs are structurally different, hub specific message types have to be defined.

In the hub oriented communication architecture scenario, two cases can be defined. In the first case, a pair of command-state topics per hub is communicating via a set of hub specific distributor-collector nodes, Fig. 3(a). The distributor nodes subscribe to the intended command topic, disassemble the message and publish matching message fragments to all nodes in the local hub. Nodes of a hub send their state to the collector node which assembles the total state of the hub in a state message and publishes it on the state topic of that hub. In this case, all the traffic to and from the hub is bridged with a single pair of one-directional static bridges.

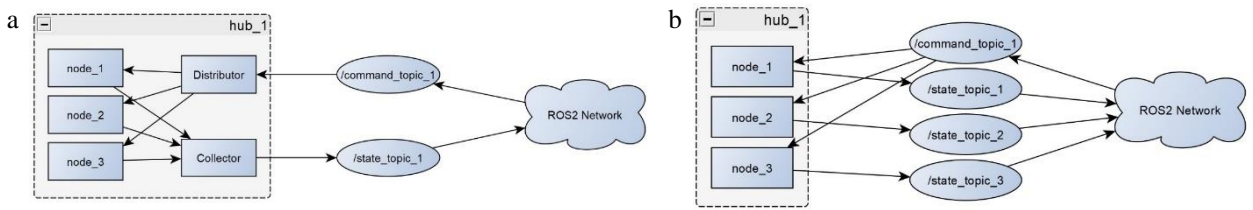

Fig. 3: (a) Distributor and collector node aided hub oriented communication; (b) Direct hub oriented communication

In the second case, there are no gateway nodes that distribute or collect messages in a hub, Fig. 3(b). This means that all nodes of a hub subscribe either to the same command topic for that hub and evaluate only a part of the message intended for them, or subscribe to node dedicated topics. Without the state collector node in a hub, each node publishes its state on a separate topic. These two cases shouldn't be considered as mutually exclusive and should be used together according to complexities of nodes, hubs and the overall system.

In a multi-hub system where hubs contain nodes that are similar to nodes from other hubs, a node oriented communication architecture can be approached. This means that similar nodes from all hubs subscribe to a single topic and evaluate the message according to some flags that specifiy the hub. In this scenario, the state can be collected in one of the ways shown in the hub oriented scenario, or by having a dedicated node that collects states from node groups from all hubs and pass it on.

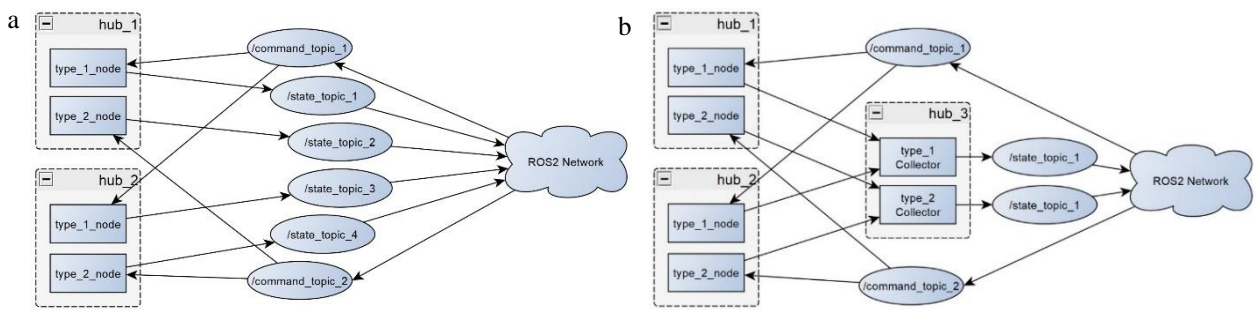

Fig. 4: (a) Direct node type oriented communication; (b) State collector hub aided node type oriented communication

In the described use-case, the communication architecture was dominantly node type oriented without the aid of a state collector hub. This means that the stream of state messages from each node occupied dedicated topics, which made the number of topics needed for communication with SP quite big. Structure was maintained utilizing adequate namespaces for each topic group.

## 6. Conclusion

It is not always trivial to come up with a good solution on how to design the system architecture in the beginning phases of a project, especially for a large and distributed automation system. This paper presented utilization of ROS and ROS2 in intelligent automation systems and pointed out ways to structure the communication architecture. A described use-case incorporated the stated proposals throughout the paper and showed possibilities of achieving flexible automation in a collaborative robot assembly station.

Some of the main features of designing a system in the described way are the ability to choose between DDS implementations and QoS policy options, an industrial standard communication protocol, the independence of a specific operating system and ROS distribution, scalability due to an architecture based on self-contained hubs and a major library of ROS and ROS2 packages containing tools and algorithms necessary for large scale intelligent automation.

Future work will include further development of stateless nodes with and attempt to make them more general and reusable. The underlying DDS layer will be additionally explored and its features utilized. Sequence Planner will be branched and a distribution for the ROS community will be provided.

## Acknowledgements

This project is supported by UNIFICATION, Vinnova, Produktion 2030.